\documentclass[runningheads]{llncs}
\usepackage[table,xcdraw]{xcolor}
\usepackage{float}
\usepackage{graphicx}
\usepackage{cite}
\begin{document}
\title{Development and Validation of a Novel Prognostic Model for Predicting AMD Progression Using Longitudinal Fundus Images}
\titlerunning{Longitudinal Deep Learning Prognostic Model}
%

\author{Joshua Bridge\inst{1} \and
Simon P. Harding\inst{1,2} \and
Yalin Zheng\inst{1,2}}
\authorrunning{J. Bridge et al.}
 %
\institute{Department of Eye and Vision Science, University of Liverpool, L7 8TX, UK \\
\email{\{joshua.bridge, s.p.harding, yalin.zheng\}@liverpool.ac.uk}\\
\url{www.liv-cria.co.uk}\and
	St. Paul’s Eye Unit, Royal Liverpool University Hospital, UK\\
	}

\maketitle              
\begin{abstract}
Prognostic models aim to predict the future course of a disease or condition and are a vital component of personalized medicine. Statistical models make use of longitudinal data to capture the temporal aspect of disease progression; however, these models require prior feature extraction. Deep learning avoids explicit feature extraction, meaning we can develop models for images where features are either unknown or impossible to quantify accurately. Previous prognostic models using deep learning with imaging data require annotation during training or only utilize a single time point. We propose a novel deep learning method to predict the progression of diseases using longitudinal imaging data with uneven time intervals, which requires no prior feature extraction. Given previous images from a patient, our method aims to predict whether the patient will progress onto the next stage of the disease. The proposed method uses InceptionV3 to produce feature vectors for each image. In order to account for uneven intervals, a novel interval scaling is proposed. Finally, a Recurrent Neural Network is used to prognosticate the disease. We demonstrate our method on a longitudinal dataset of color fundus images from 4903 eyes with age-related macular degeneration (AMD), taken from the Age-Related Eye Disease Study, to predict progression to late AMD.  Our method attains a testing sensitivity of 0.878, a specificity of 0.887, and an area under the receiver operating characteristic of 0.950. We compare our method to previous methods, displaying superior performance in our model. Class activation maps display how the network reaches the final decision.

\end{abstract}

\section{Introduction}
Prognostic models are an essential component of personalized medicine, allowing health experts to predict the future course of disease in individual patients \cite{steyerberg}. Advances in computing power and an abundance of data have allowed for increasingly sophisticated models to be developed. Most developed prognostic models use statistical methods such as logistic regression; these models require prior feature extraction, either manual or automatic \cite{manual} and are limited in the number of included variables. Feature extraction can be costly and time-consuming, especially in imaging data. Deep learning offers the ability to avoid explicit feature extraction, allowing us to develop models without the need for handcrafted features. For this reason, deep learning is especially useful in imaging data. Prognostic deep learning models have been developed in several fields, primarily ophthalmology \cite{sisternes}, cardiology \cite{kwon}, and neurology \cite{hilario}, and several modalities, including magnetic resonance imaging (MRI), optical coherance tomography (OCT), color fundus photography and X-Ray.
\par Current prognostic models that utilize deep learning to analyze imaging data, either use automatic feature extraction algorithms to extract known features or only consider a single time point. Models developed using feature extraction, train algorithms on annotated images to extract relevant features such as volumes in OCT data; those features are then fed into a traditional statistical model, see \cite{sisternes, niu, leng} for examples. Manual feature extraction is time-consuming and requires expert readers. More recently, Yim et al. \cite{yim} proposed a method which automatically segments OCT layers before classification. This method outperformed human experts; however, automatic feature extraction requires annotations during training, which is not always available in situations when the features are unknown or difficult to quantify, such as is the case when using color fundus imaging. 
\par An alternative to explicit feature extraction is to use deep learning to extract features implicitly, such as used by \cite{babenko, arcadu}. Many models take the previous available image and fit a pretrained convolutional neural network (CNN), with Inception V3 \cite{iv3} being a popular choice due to its generalizability and high performance in a variety of tasks. This method, unlike the feature extraction method, may be applied to any image even when features are not explicitly known; however, this creates a separate issue, by using only one image, these models may fail to capture the temporal pattern across time points.
\par Here, we develop a prognostic model to predict the progression of disease, from longitudinal images. The method is applicable to any modality even when the causes of progression are unknown or can't be quanitfied. The proposed method is demonstrated on a dataset consisting of 4903 eyes with age-related macular degeneration (AMD), taken from the AREDS dataset \cite{areds1}. The method is generalizable to any longitudinal imaging data. We show that by considering the time interval between images and adopting a method from time series analysis, we can provide significantly improved prediction performance.

Our contributions are as follows:
\begin{itemize}
    \item Propose a novel method to predict the future prognosis of a patient from longitudinal images
    \item Introduce interval scaling which allows for uneven time intervals between visits
    \item Demonstrate on the largest longitudinal dataset and attain state-of-the-art performance outperforming other state-of-the-art methods
\end{itemize}

\section{Method}
Given images $ \{ X_0, \dots, X_i, \dots, X_N \}$ at times $\{t_0, \dots, t_i, \dots, t_N\}$, we wish to predict the diagnosis $y_{N+1}$ at time $t_{N+1}$, where $t_{i+1} - t_i = t_i  - t_{i-1}$ does not necessarily hold, which is common in a clinical setting.
\par
\begin{figure}[H]
\centering
\includegraphics[width=.7\textwidth]{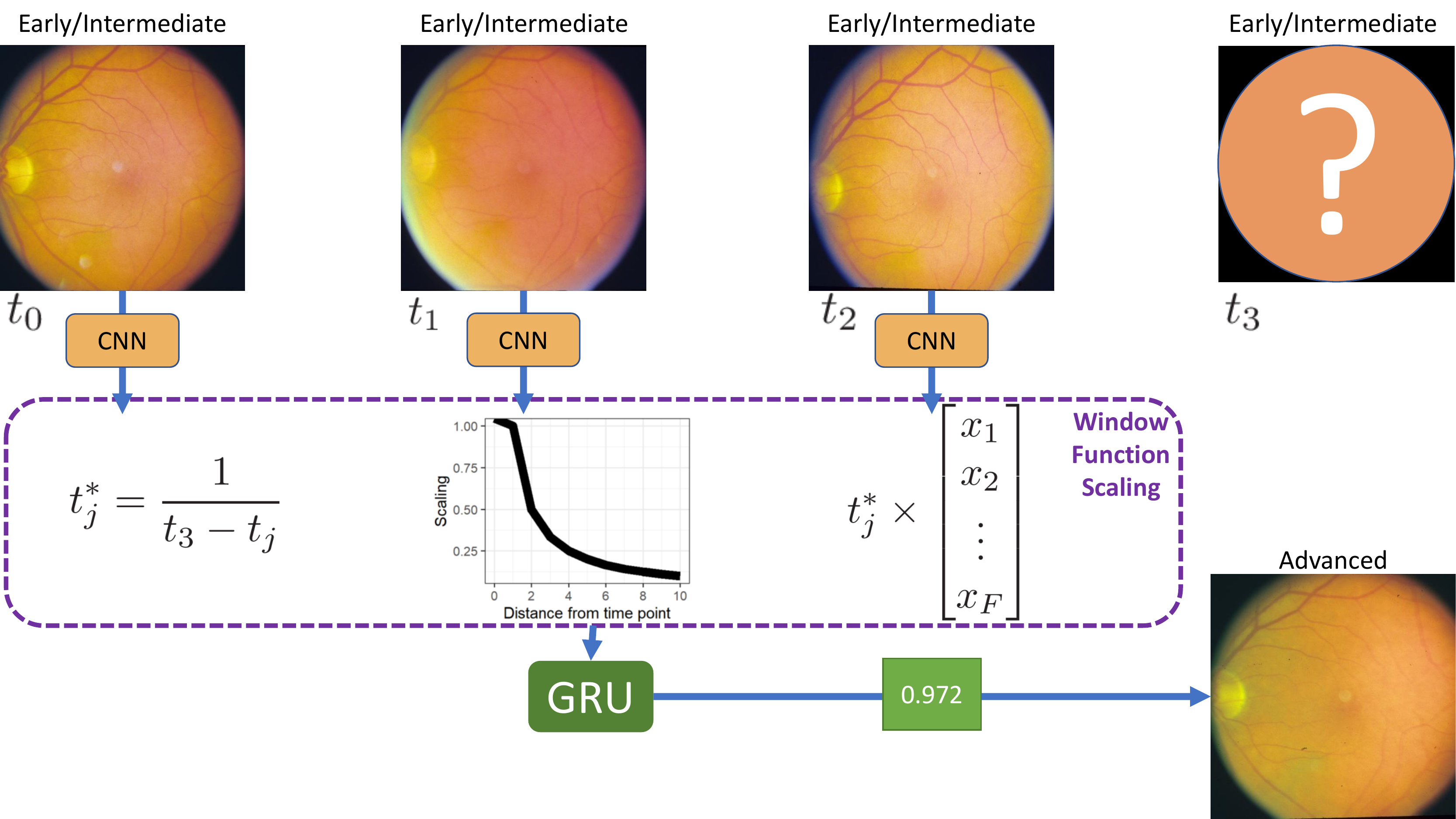}
\caption{Overview of the proposed method. For each of the T time points, we fit a CNN with shared weights, resulting in a vector of length F, per image. Each vector is multiplied by a corresponding interval scaling. The scaled vectors are combined into a single T$\times$F matrix and a Gated Recurrent Unit (GRU) with sigmoid activation gives a probability of progression. For simplicity, 3 time points are shown, this method is extendable to any number of time points.}
\label{fig:framework}
\end{figure}

 The proposed method consists of three stages, firstly, we utilize a pretrained CNN, with shared weights, to reduce each image to a single feature vector. Then, the feature vectors are combined, and an interval scaling is applied to account for the uneven time intervals, this weights the most recent time points as being more important in making the final prediction. Finally, a recurrent neural network (RNN) classifies the images as progressing or non-progressing. An overview of the proposed framework is shown in Figure \ref{fig:framework}.

\subsection{Inception V3}
We begin by fine-tuning a pretrained CNN on each image, with shared weights, to extract feature vectors. In our work, we chose IncpetionV3 \cite{iv3} pretrained on ImageNet \cite{imagenet}. InceptionV3 increases accuracy over previous networks, while remaining computationally efficient, through the use of factorized kernels, batch normalization, and regularization. InceptionV3 is considered highly generalizable with a greater than 78.1\% accuracy on the ImageNet dataset.  The InceptionV3 network results in a feature vector of length F=2048 for each image at each time point. This network has previously been used to provide state of the art results in single time point methods \cite{babenko, arcadu}, and is used here as a feature extractor.

\subsection{Interval scaling}
To account for uneven time intervals, we implement a triangular window function to create a smoothing model. Whereas in a simple moving average model, the time points are weighted equally, smoothing models weight values closer to $t_{N+1}$ as being more useful in the prediction. For each sequence of images at times, $t_0, t_1, \dots, t_i, \dots, t_{N+1}$, where $t_{N+1}$  is the time point that we want to predict at, we rescale each time such that $t_i^* = 1/(t_{N+1} - t_i)$. The feature vectors for each image are then multiplied by their corresponding time interval scale. This scaling weights the images such that images closer to the time point of interest are considered more important than those observed at further time points, thus allowing the network to account for uneven time intervals.

\section{GRU prediction}
To predict whether the patient will progress to advanced AMD or not, we combine the interval corrected vectors into a $T\times F$ matrix, where $T$ corresponds to the number of time points and $F$ is the number of features. We apply a Gated Recurrent Unit (GRU) \cite{gru1} with a filter size of 1, resulting in a single value. GRU was chosen as opposed to Long Short-Term Memory (LSTM) \cite{lstm} units, as GRU is more computationally efficient. LSTM units perform better on longer sequences; however, in this case, we only have three time points \cite{gru2}. The sigmoid activation function then scales the value between 0 and 1. Any values greater than a threshold of 0.5 are predicted to be future progressing patients.

\section{Experiments}
In order to evaluate the performance, the proposed method is demonstrated on a dataset of AMD images with two and three time points and compared to a single time point method.
\subsection{Data}
Data consists of color fundus images taken from the Age-Related Eye Disease Study (AREDS) \cite{areds1}, the most extensive clinical study into AMD. 
\par AMD is a leading cause of vision loss worldwide \cite{amdworld}. There are two main stages of AMD, early/intermediate, defined by small- to medium-sized drusen, and advanced, defined by geographic atrophy (GA) or neovascularization (nAMD) \cite{areds1}. Drusen can be observed as yellow-white lipid deposits under the retina, varying greatly in size and morphology \cite{drusen}. The exact causes of AMD are unknown; however, studies have shown that smoking and genetics are significant risk factors \cite{factors}. Risk factors for progression from early/intermediate to advanced AMD are also unknown; however, there is evidence that drusen and optic disk characteristics are important \cite{disc1, disc2}. Vision loss can be avoided with interventions such as anti-VEGF treatment; however, disease progression and the need for treatment are often hard to predict \cite{vegf}. This highlights the need for accurate prognostic models. 
\par We extracted 4,903 eyes, which had four visits, complete with images and diagnoses at each visit, with no diagnosis of advanced AMD during the first three visits. Advanced AMD was defined as either Central GA, nAMD, or both GA and nAMD. We used the last visit as ground truth to make our prediction based on the first three visits. Of the 4,903 included eyes, 453 (9.2\%) progressed to advanced AMD. 
\par
\begin{figure}[H]
\centering
\begin{tabular}{cccc}
  \includegraphics[width=20mm]{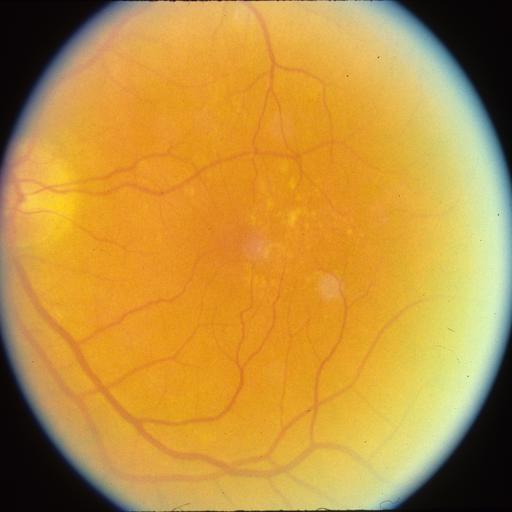} &   \includegraphics[width=20mm]{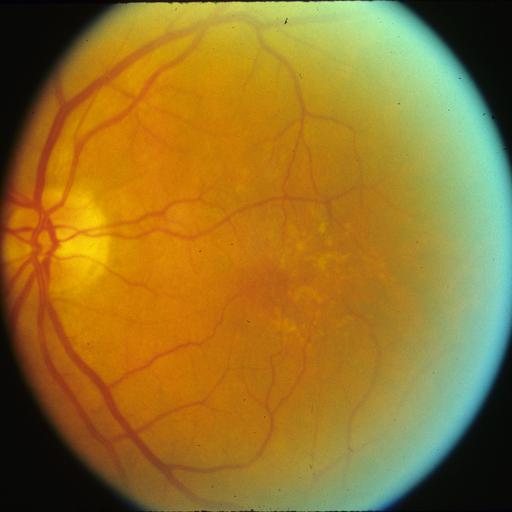} &   \includegraphics[width=20mm]{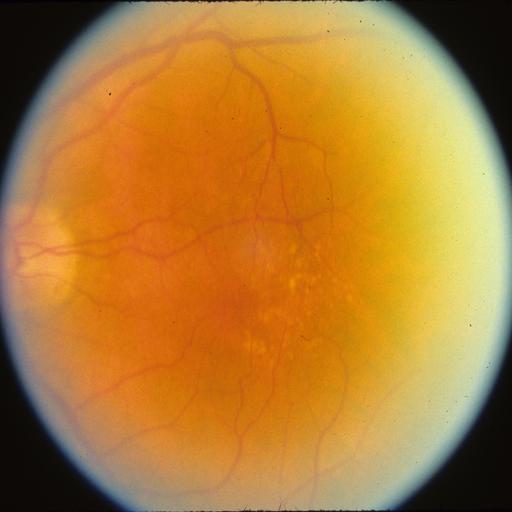} &   \includegraphics[width=20mm]{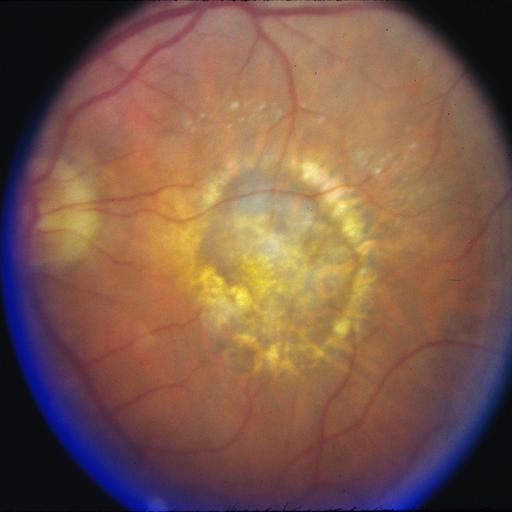} \\
  0 years &  2 years & 3 years & 7.5 years\\
  Early/intermediate &  Early/intermediate & Early/intermediate & Advanced\\
\multicolumn{4}{c}{(a) Progressing patient}  \\[6pt]
  \includegraphics[width=20mm]{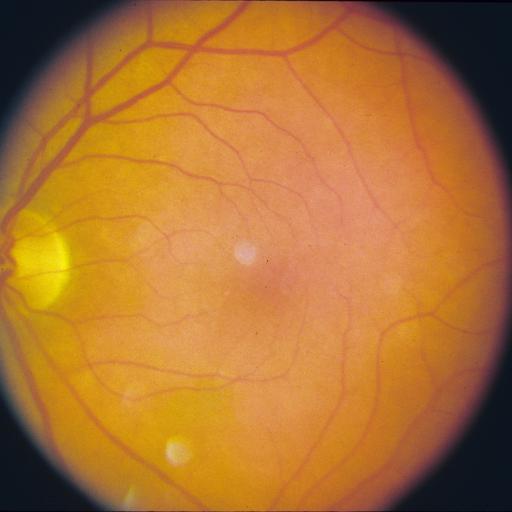} &   \includegraphics[width=20mm]{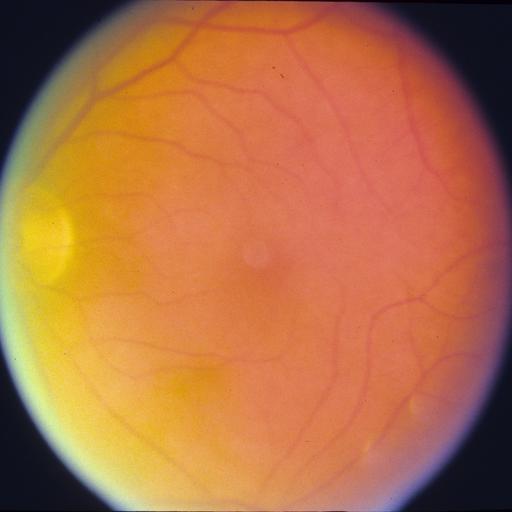} &   \includegraphics[width=20mm]{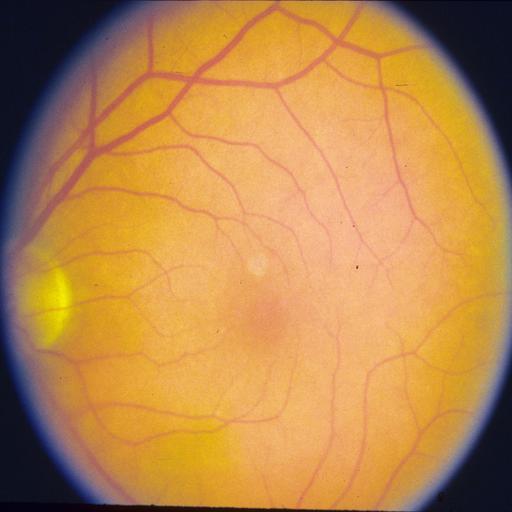} &   \includegraphics[width=20mm]{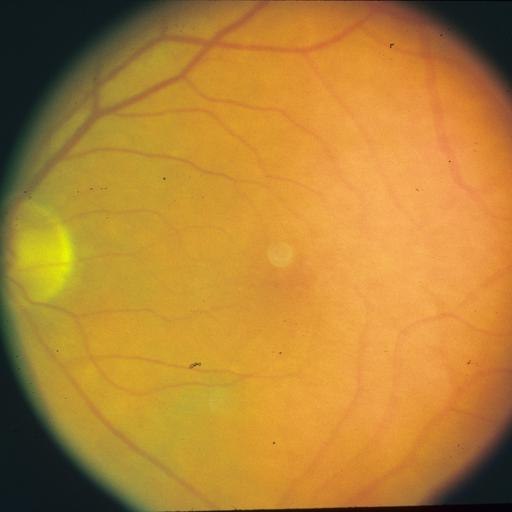} \\
  0 years &  2 years & 3 years & 4 years\\
  Early/intermediate &  Early/intermediate & Early/intermediate & Early/intermediate\\
\multicolumn{4}{c}{(b) Non-progressing patient }  \\[6pt]
\end{tabular}

\caption{Sample images from a progressing patient (top) and non-progressing patient (bottom). The first three images show early/intermediate AMD, while the fourth image shows whether they progressed to advanced AMD or not.}

\label{fig:images}
\end{figure}

 We randomly split the data into 60\% training (2942 eyes, 272 progressing), 20\% validation (981 eyes, 91 progressing), and 20\% testing (980 eyes, 90 progressing) datasets. To reduce the possibility of data leakage, patients with both eyes included were kept within the same data split. Example images are given in Figure \ref{fig:images}.

\subsection{Preprocessing}
Any images in the dataset where the patient had already progressed to advanced AMD, or without the required three previous images plus a fourth prediction image for prediction, were excluded. The images were automatically cropped by first calculating the difference between the original image and the background color, an offset was added to the difference, and the bounding box was calculated from this. Image values were rescaled from between 0 and 255, to between 0 and 1. All images were resized to 256x256 pixels to reduce computational requirements. Right eye images were flipped, such that the optic disc on all images was located on the left. No prior feature extraction or segmentation/registration is required, such that our method is as generalizable to other diseases and modalities as possible. All preprocessing was automated, with no subjective human input required.

\subsection{Computing}
All analyses were carried out on a Linux machine with a Titan X 12GB GPU and 32GB of memory. Deep learning was conducted in Python 3.7  using the Keras 2.2.4 library \cite{keras} with TensorFlow \cite{tf} as the base library. Confidence intervals were calculated using R 3.4.4 \cite{r}, with the pROC package \cite{proc}. 
\par Optimization was carried out with the Adam optimizer \cite{adam} with an initial learning rate of 0.0001. We used binary cross-entropy as the loss function. If the loss did not improve after ten epochs, then the learning rate was reduced to two-thirds. Model checkpoints and early stopping prevented overfitting, with the best model being picked according to the validation loss.

\subsection{Metrics}
We evaluate model performance using the commonly used area under the receiver operating characteristic curve (AUC) \cite{auc}, optimal sensitivity, and optimal specificity, determined by Youden's index. To assess whether the difference in these measures between models is significant, we construct confidence intervals. De Long's method \cite{delong} is used to construct confidence intervals for AUC, and bootstrapping with 2000 samples is used for sensitivity and specificity to calculate 95\% confidence intervals. Results from De Long's test \cite{delong} are also reported.

\subsection{Results and comparisons}
Results are reported using two and three time points with our method, to assess the benefit of adding additional time points. We compare our results with a method similar to those used in previous work \cite{babenko, arcadu}, using single time points with a CNN. Taking the last available image, we fine tune InceptionV3 \cite{iv3} pretrained on ImageNet \cite{imagenet} to classify as progression or no progression.  

\par The proposed method using three time points achieves an AUC,  optimal sensitivity, and optimal specificity of 0.950 (0.923, 0.977), 0.878 (0.810, 0.945), and 0.887 (0.866, 0.907), respectively; this is a significant improvement over the single time point method which had AUC, sensitivity, and specificity of 0.857 (0.823, 0.890), 0.867 (0.796, 0.937), and 0.760 (0.731, 0.788). These results show a statistically significant increase in AUC and specificity and a non-significant increase in sensitivity when using the proposed three time point method over the previous single time point methods. De Long's test gave a p-value  $<$ 0.0001, indicating a significant difference in AUCs. This significant increase in specificity without a loss in sensitivity shows our model can reduce false positives without increasing false negatives, over the previous model.

\par
\begin{figure}[H]
\centering
\includegraphics[width=0.75\textwidth]{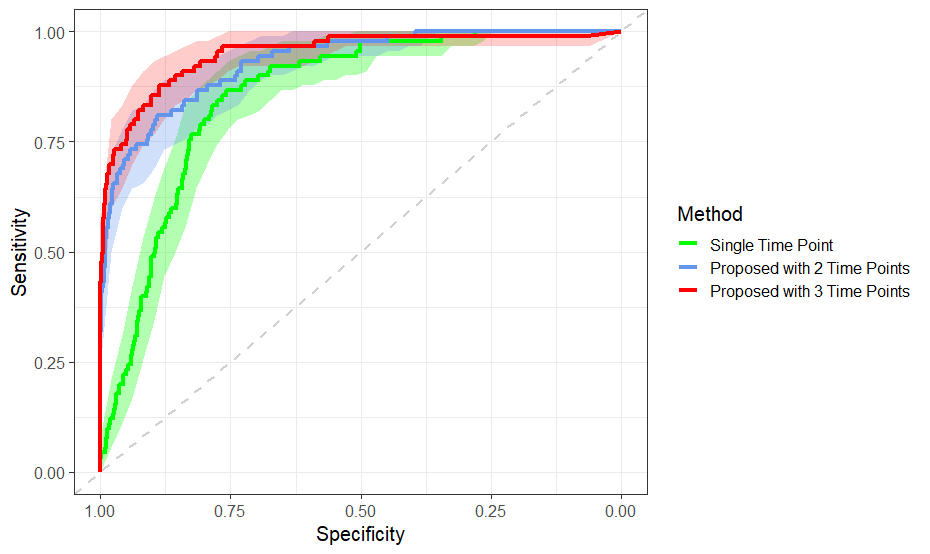}
\caption{Receiver operating characteristic curve for the single time point InceptionV3 method, the proposed method with 2 time points, and the proposed time point with three time points. Increasing the number of time points appears to increase the area under the curve. Faded bands show 95\% confidence intervals.}
\label{fig:roc}
\end{figure}

 The method utilizing two time points gave an AUC, sensitivity, and specificity of 0.932 (0.905, 0.958), 0.811 (0.730, 0.892), 0.760 (0.731, 0.788). The three time point method had a non-significant increase over two time points. This may indicate that in this example more than two time points does not add any significant predictive value. Results are presented in Table \ref{tab:results} and the receiver operating characteristic is shown in Figure \ref{fig:roc}. Experiments without interval scaling were also conducted and showed a significant decrease in performance.

\begin{table}
\begin{center}
\caption{Area Under the Receiver Operating Characteristic (AUC) with $95\%$ confidence intervals constructed by De Long's method. Sensitivity and Specificity with $95\%$ confidence intervals constructed by bootstrapping with $2000$ samples.}
\label{tab:results}
\begin{tabular}{c|c|c|c}
\hline
                                                                     & AUC                                                             & Sensitvity                                                      & Specificity                                                     \\ \hline
\begin{tabular}[c]{@{}c@{}}Single image\\ method\end{tabular}        & \begin{tabular}[c]{@{}c@{}}0.857 \\ (0.823, 0.890)\end{tabular} & \begin{tabular}[c]{@{}c@{}}0.867 \\ (0.796, 0.937)\end{tabular} & \begin{tabular}[c]{@{}c@{}}0.760 \\ (0.731, 0.788)\end{tabular} \\ \hline
\begin{tabular}[c]{@{}c@{}}Proposed method\\ (2 images)\end{tabular} & \begin{tabular}[c]{@{}c@{}}0.932 \\ (0.905, 0.958)\end{tabular} & \begin{tabular}[c]{@{}c@{}}0.811 \\ (0.730, 0.892)\end{tabular} & \begin{tabular}[c]{@{}c@{}}0.892 \\ (0.872, 0.913)\end{tabular} \\ \hline
\begin{tabular}[c]{@{}c@{}}\textbf{Proposed method}\\ \textbf{(3 images)}\end{tabular} & \begin{tabular}[c]{@{}c@{}}\textbf{0.950} \\ \textbf{(0.923, 0.977)}\end{tabular} & \begin{tabular}[c]{@{}c@{}}\textbf{0.878} \\ \textbf{(0.810, 0.945})\end{tabular} & \begin{tabular}[c]{@{}c@{}}\textbf{0.887} \\ \textbf{(0.866, 0.907)}\end{tabular} \\ \hline
\end{tabular}
\end{center}
\end{table}

\subsection{Class activation maps}
\begin{figure}
\centering
\begin{tabular}{cccccc}
  \includegraphics[width=18mm]{1468_RE_t3_00_ch.jpg} &   \includegraphics[width=18mm]{1468_RE_t2_04_ch.jpg} &   \includegraphics[width=18mm]{1468_RE_t1_06_ch.jpg} &  \includegraphics[width=18mm]{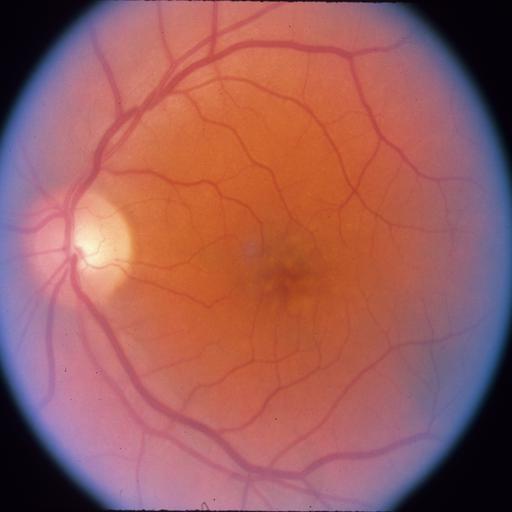} &   \includegraphics[width=18mm]{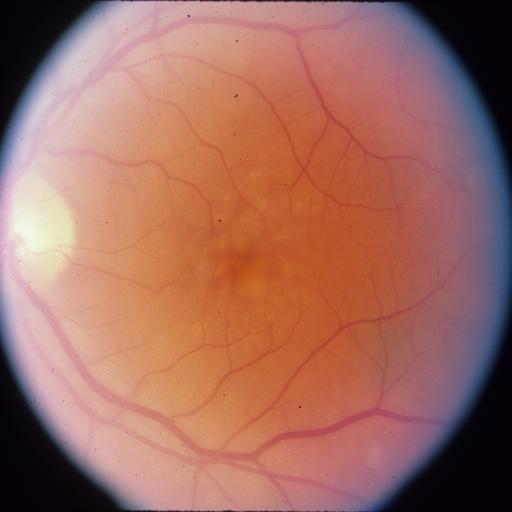} &   \includegraphics[width=18mm]{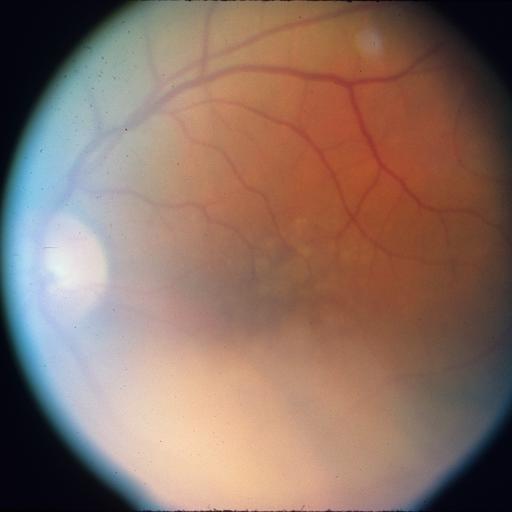} \\
  \includegraphics[width=18mm]{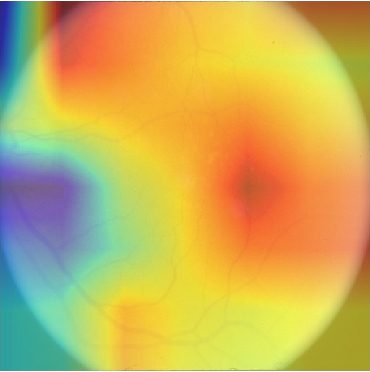} &   \includegraphics[width=18mm]{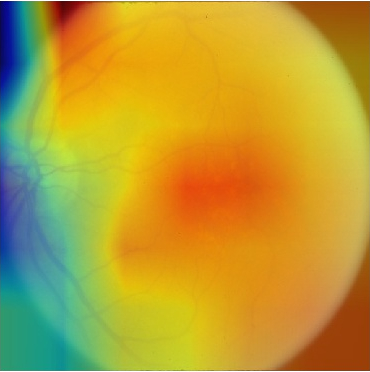} &   \includegraphics[width=18mm]{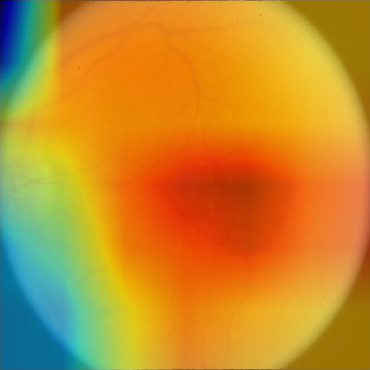} &  \includegraphics[width=18mm]{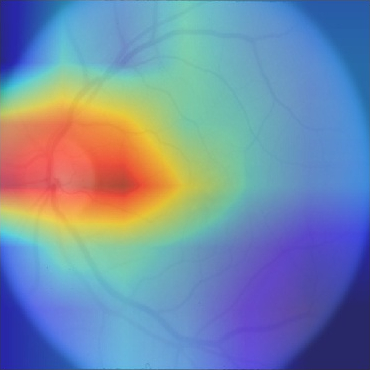} &   \includegraphics[width=18mm]{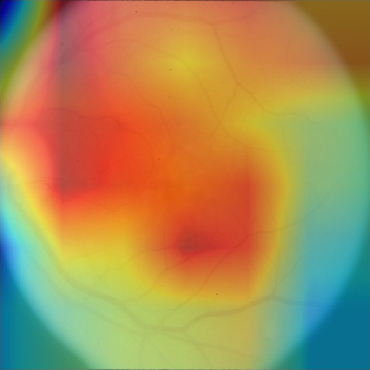} &   \includegraphics[width=18mm]{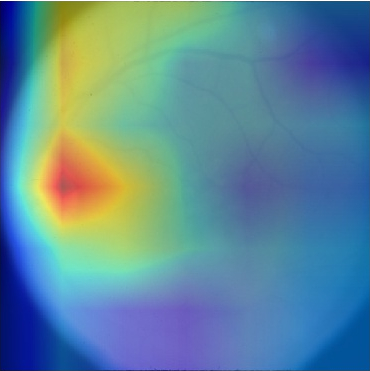} \\
  0 years & 2 years & 3 years & 0 years & 2 years & 3 years\\
\multicolumn{3}{c}{Eye 1} & \multicolumn{3}{c}{Eye 2} \\
\end{tabular}
\label{fig:cam}
\caption{Class activation maps show the areas that the network finds useful in making the prediction. Original images are also shown for reference. The network correctly identifies areas of interest in AMD progression. In blurred images, drusen are difficult to see; the class activations show that the network uses the optic disk in to reach decisions in this case. All example images are taken from the testing dataset.}
\end{figure}

To determine if our network is identifying the correct features and to reduce the black-box nature of deep learning, we create class activation maps \cite{cam} for each time point. We altered the top of the network slightly to achieve this, adding a dense layer after the GRU layer. While this altered network showed no significant change in predictive performance, it increased the network size by around a factor of 2. The class activation maps are shown in Figure \ref{fig:cam}, alongside original images for comparison.  
\par The class activation maps show that areas with high concentrations of drusen are considered relevant by the network; this is expected and shows that our network is identifying the correct features. In some images, the optic disk is also highlighted, confirming that optic disk characteristics are indeed important factors in AMD progression, as observed previously \cite{disc1, disc2}. In images where drusen are challenging to see, the network appears to use the optic disk solely in making a prediction. It is also interesting to note that the network seems to be surer of the area of interest in images that are closer to the prediction time point. In a clinical setting, these maps may are useful when justifying the prediction.

\section{Conclusions}
In this work, we proposed a novel deep learning prognostic model to predict the future onset of disease. The proposed method addresses the challenge of analyzing multiple longitudinal images with uneven time points, without the need for prior image annotation. Introducing an interval scaling was shown to improve performance over a single time point method significantly. We show that by taking into account the varying times between observed images, we can significantly improve the performance of a longitudinal prognostic model. Our method provides a statistically significant increase in specificity, which is critical in contexts such as screening. Our method utilizes time intervals meaning we can extend the interval to the observed outcome to predict further into the future; this is useful in a screening context. Future work is required to assess the generalizability of the proposed method to other diseases and to extend its use to a screening context.

\bibliographystyle{splncs04}
\bibliography{bib_miccai}
\end{document}